%% file: arxiv.tex
\documentclass[10pt,twocolumn,letterpaper]{article}

\usepackage{cvpr}

\usepackage{float}
\usepackage{caption}
\usepackage{booktabs}
\usepackage[utf8]{inputenc}
\usepackage[T1]{fontenc}
\usepackage{url}
\usepackage{amsfonts}
\usepackage{amsthm}
\usepackage{amsmath}
\usepackage{nicefrac}
\usepackage{microtype}
\usepackage{xcolor}
\usepackage{graphicx}
\usepackage{multirow}

\definecolor{cvprblue}{rgb}{0.21,0.49,0.74}
\newtheorem{theorem}{Theorem}[section]

\input{sections/math_commands}

\usepackage[pagebackref,breaklinks,colorlinks,allcolors=cvprblue]{hyperref}

\title{ESCAPE: Equivariant Shape Completion via Anchor Point Encoding}

\author{
  Burak Bekci\textsuperscript{1}, Nassir Navab\textsuperscript{1}, Federico Tombari\textsuperscript{1,2}, Mahdi Saleh\textsuperscript{1} \\
  \textsuperscript{1}Technical University Munich (TUM), \textsuperscript{2}Google\\
  \texttt{\{burak.bekci, navab, m.saleh\}@tum.de, tombari@in.tum.de}
}

\begin{document}

\maketitle

\begin{abstract}
Shape completion, a crucial task in 3D computer vision, involves predicting and filling the missing regions of scanned or partially observed objects. Current methods expect known pose or canonical coordinates and do not perform well under varying rotations, limiting their real-world applicability. We introduce \textbf{ESCAPE} (Equivariant Shape Completion via Anchor Point Encoding), a novel framework designed to achieve rotation-equivariant shape completion. Our approach employs a distinctive encoding strategy by selecting anchor points from a shape and representing all points as a distance to all anchor points. This enables the model to capture a consistent, rotation-equivariant understanding of the object’s geometry. ESCAPE leverages a transformer architecture to encode and decode the distance transformations, ensuring that generated shape completions remain accurate and equivariant under rotational transformations. Subsequently, we perform optimization to calculate the predicted shapes from the encodings. Experimental evaluations demonstrate that ESCAPE achieves robust, high-quality reconstructions across arbitrary rotations and translations, showcasing its effectiveness in real-world applications without additional pose estimation modules.
\end{abstract}

\input{sections/01_intro}
\input{sections/02_related}
\input{sections/03_method}
\input{sections/04_experiments}

\input{sections/05_conclusion}

{
\small
\bibliographystyle{splncs04}
\bibliography{refs}
}
\input{sections/supplementary}


\end{document}

%% file: sections/math_commands.tex
\usepackage{amsmath,amsfonts,bm}

\def\eqref#1{equation~\ref{#1}}

\def\1{\bm{1}}

\DeclareMathAlphabet{\mathsfit}{\encodingdefault}{\sfdefault}{m}{sl}
\SetMathAlphabet{\mathsfit}{bold}{\encodingdefault}{\sfdefault}{bx}{n}



%% file: sections/01_intro.tex
\section{Introduction}
3D perception has been dependent on canonical orientations, causing significant challenges in dynamic environments, such as robotic manipulation or real-time object recognition, where objects interact from varying viewpoints. Traditional methods, like voxel-based approaches and early point cloud networks such as PointNet\cite{qi2017pointnet} and PointNet++\cite{qi2017pointnet++}, have laid a strong foundation but cannot inherently handle rotational variance. Recent methods have introduced improvements by leveraging attention mechanisms and hierarchical processing capabilities but still rely on orientation normalization or data augmentation techniques to perform robustly with rotational changes, which do not fully resolve the inherent challenges of varying viewpoints.

\begin{figure}[H]
\centering
\includegraphics[width=\linewidth]{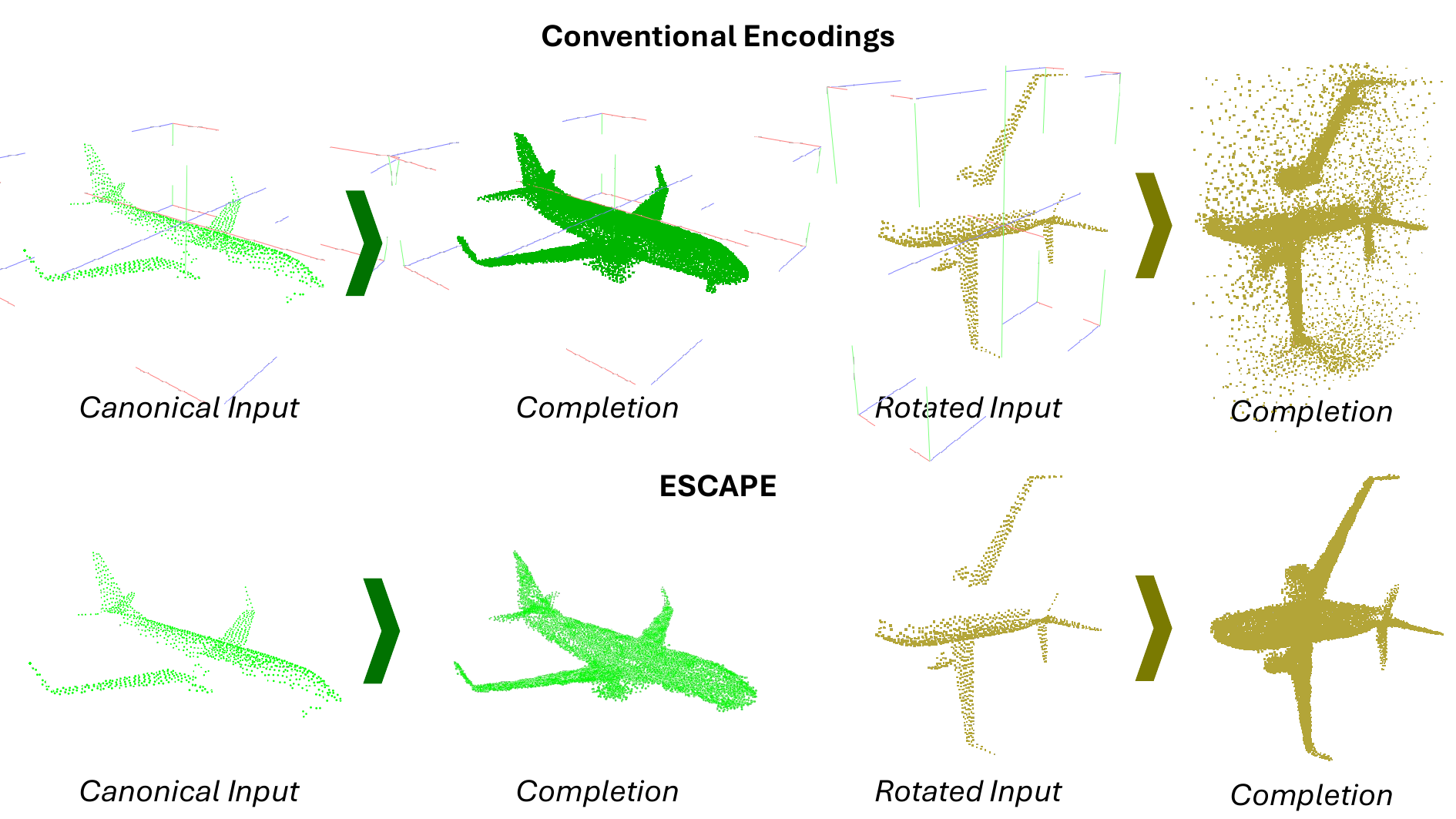}
\caption{Existing shape completion methods (Top) use conventional canonical coordinates and perform poorly under rotation changes and unknown canonical reference. Using our anchor point encoding (Below), we manage to consistently complete the shape with arbitrary rotation in non-canonical coordinates. }
\label{fig:teaser}
\end{figure}

Despite progress in the field, existing shape completion techniques struggle to maintain robustness and accuracy under arbitrary rotations. This notably limits their practicality in real-world applications where objects can appear in any orientation. Recent methods, including transformer-based models like SnowflakeNet~\cite{xiang2021snowflakenet} and SeedFormer~\cite{zhou2022seedformer}, have introduced improvements by leveraging the attention mechanisms, progressively growing points, and hierarchical processing capabilities of transformers. However, these models expect partial shapes in aligned orientation across a category and only learn completion in canonical coordinates, which do not guarantee performance consistency in varying viewpoints or unknown local coordinates.

Recent advances in rotation-invariant descriptors, particularly through Vector Neurons and SE(3)-equivariant networks, have shown promise in tasks like registration and classification. However, these approaches face limitations when applied to shape completion: their layerwise transformations can compound errors through deep networks ($\mathcal{O}(\alpha^L)$ for $L$ layers), and their equivariant constraints may sacrifice geometric information critical for completing missing regions. While these methods achieve equivariance, they don't provide guarantees on geometric completeness - two shapes with identical vector features aren't necessarily isometric.

To address these challenges, we introduce ESCAPE, a novel framework specifically designed for rotation-equivariant shape completion. Our key insight is representing shapes through distances to dynamically selected anchor points, which provides several theoretical advantages: constant error bounds ($\mathcal{O}(1)$) independent of network depth, guaranteed geometric completeness (identical distance encodings imply isometric shapes), and provable reconstruction uniqueness with sufficient anchor points. ESCAPE processes this representation through a transformer architecture, leveraging attention mechanisms to predict distances to anchor points for the complete shape while maintaining rotational equivariance. This approach not only preserves geometric fidelity crucial for completion but also offers practical benefits - simpler optimization without specialized equivariant layers and natural handling of varying partiality through adaptive anchor selection.

We propose a complete pipeline to generate, encode, decode, and interpret rotation equivariant features for non-canonical 3D shape completion. 

In summary, our contributions are threefold:
\begin{itemize}
\item We introduce ESCAPE, a novel rotation-equivariant 3D encoding strategy using high-curvature anchor points, enabling robust shape description and reconstruction.
\item We develop a transformer-based architecture that leverages our equivariant encoding to generate the completed point cloud, maintaining consistency across varying object orientations and partial inputs.
\item We present the an end-to-end rotation-equivariant shape completion method, demonstrating robust performance under arbitrary rotations and in the absence of canonical object coordinates.
\item We establish and evaluate a challenging real-world shape completion benchmark using the OmniObject dataset\cite{wu2023omniobject3d}, featuring partial point clouds with diverse geometries and arbitrary poses, bringing shape completion works to real-world data.
\end{itemize}

%% file: sections/02_related.tex
\section{Related Works}
\subsection{Point Cloud Processing}

Point clouds, representing unordered and sparse 3D data, challenge conventional CNNs. Voxel-based methods like VoxNet~\cite{maturana2015voxnet} and VoxelNet~\cite{zhou2018voxelnet} convert point clouds into regular grids for 3D CNN processing but are computationally intensive. PointNet~\cite{qi2017pointnet} and PointNet++~\cite{qi2017pointnet++} address unordered sets with MLPs and hierarchical grouping. PointNet creates a permutation-invariant structure with pointwise MLPs, while PointNet++ adds hierarchical layers and furthest point sampling, increasing computational complexity. These methods are often combined with voxel-based approaches in large-scale applications~\cite{shi2020pv,zhou2018voxelnet}.

Graph neural networks (GNNs) like PointCNN~\cite{li2018pointcnn} and EdgeConv~\cite{wang2019dynamic} capture local geometric features through spatial relationships and dynamic graph construction. Recent advancements integrate edge-based attention for improved efficiency~\cite{velivckovic2017graph, vaswani2017attention}. Transformers and attention mechanisms have also been applied to point clouds, with Graph Attention Networks (GATs)~\cite{velivckovic2017graph} using self-attentional layers and the Transformer architecture~\cite{vaswani2017attention} introducing multi-head attention for robust feature aggregation. Set Transformer~\cite{lee2019set} and Perceiver~\cite{jaegle2021perceiver} adapt attention mechanisms for 3D data. Point cloud-specific transformers, such as PCT~\cite{guo2020pct} and Point Transformer~\cite{zhao2021point}, apply self-attention to neighborhood points, though they can be computationally expensive. 

\subsection{Rotation-Invariant Encodings}
Handcrafted rotation-invariant descriptors have been widely explored in 3D by researchers before the popularity of deep neural networks. To guarantee invariance under rotations, many handcrafted local descriptors~\cite{rusu2008aligning,rusu2009fast,tombari2010unique,guo2013rotational} rely on an estimated local reference frame (LRF), which is typically based on the covariance analysis of the local surface, to transform local patches to a canonical representation. The major drawback of LRF is its non-uniqueness, making the constructed rotational invariance fragile and sensitive to noise. Consequently, attention has shifted to LRF-free approaches~\cite{drost2010model}. These methods focus on mining the rotation-invariant components of local surfaces to represent the local geometry. For instance, PPF~\cite{drost2010model}, PFH~\cite{rusu2008aligning}, and FPFH~\cite{rusu2009fast} encode the geometry of the local surface using histograms of pairwise geometrical properties. Early work by Osada et al.~\cite{osada2002shape} introduced shape distributions using distances between random surface points as rotation-invariant descriptors. Despite being rotation-invariant by design, these handcrafted descriptors are often inadequate for complex geometry and noisy data.

Recently, many deep learning-based methods have aimed to learn rotation-invariant descriptors. PPF-FoldNet~\cite{deng2018ppf} encodes PPF patches into embeddings, using a FoldingNet~\cite{yang2018foldingnet} decoder to reconstruct the input, enabling correspondences from the rotation-invariant embeddings. SpinNet~\cite{ao2021spinnet} and Graphite~\cite{saleh2020graphite,saleh2022bending} align local patches to defined axes before learning descriptors. However, these methods are limited by their locality, as descriptors are learned only from the local region, making them less distinctive.

YOHO~\cite{xu2021you} introduces a rotation-equivariant approach by leveraging an icosahedral group to learn a group of rotation-equivariant descriptors for each point. Rotational invariance is achieved by max-pooling over the group, but this method struggles with efficiency and complete rotational coverage. Object-centric registration methods~\cite{yew2020rpm,fu2021robust} strengthen rotational invariance by combining rotation-invariant descriptors with rotation-variant inputs, though performance drops under large rotations.

In point cloud classification, methods~\cite{zhang2019rotation,deng2021vector} describe whole shapes as rotation-invariant descriptors but lack global awareness in node/point descriptors. Methods such as Transformer-based approaches~\cite{vaswani2017attention} aim to incorporate global context, but robustness to rotational changes is often achieved through data augmentation, which is not optimal. Recent works such as Rotation-Invariant Transformer~\cite{yu2023rotation} and Riga~\cite{yu2024riga} have proposed more robust rotation-invariant and globally-aware descriptors. Despite these advancements, these approaches have not yet been thoroughly explored in shape completion tasks, highlighting a significant area for future research.

More recently, several approaches have specifically addressed equivariant shape reconstruction and completion. Vector Neurons~\cite{deng2021vector} and their adaptations~\cite{wu2022so} provide a general SO(3)-equivariant framework, while methods like SE(3)-Equivariant Networks~\cite{chen20223d} and SCARP~\cite{sen2023scarp} focus on reconstruction and completion tasks. However, these approaches face fundamental limitations: error accumulation through deep networks and potential loss of geometric information due to equivariant constraints. Our work differs by providing provable geometric completeness and constant error bounds through distance-based representation, while specifically modeled for shape completion challenges.

\subsection{Shape Completion Methods}

Pre-deep learning point cloud completion methods rely on object symmetry~\cite{mitra2013symmetry} or a database of complete shapes~\cite{pauly2005example} to achieve effective results. However, these methods are constrained by the need for specific preconditions to be met by the input data, limiting their applicability compared to deep learning-based approaches.

Learning-based methods can be classified into two primary categories based on the type of representation used for 3D data: methods utilizing point clouds and methods employing alternative representations such as voxel grids, implicit functions, and others. Methods using alternative representations often need higher memory consumption, making them less scalable for high-resolution inputs. Although researchers have proposed more efficient representations such as Octrees~\cite{wang2017cnn}, sparse lattice networks~\cite{rosu2019latticenet}, and sparse convolution operations~\cite{graham20183d}, none have proven to be as efficient and effective as directly processing 3D point coordinates.

PointNet~\cite{qi2017pointnet} and its variant PointNet++~\cite{qi2017pointnet++} introduced specialized operations that enable learning directly from point cloud coordinates, revolutionizing point cloud processing tasks. PCN~\cite{yuan2018pcn} was built on top of PointNet and became the first deep learning-based method for point cloud completion, utilizing an encoder-decoder architecture with a folding operation to complete given partial input point clouds. Following PCN, many other methods with similar model architectures have been developed~\cite{yuan2018pcn}. In addition to these methods, some researchers formulate point cloud completion as a probabilistic problem, where a given input cloud can be mapped to multiple complete point clouds. They address point cloud completion by introducing probabilistic methods such as Variational Autoencoders~\cite{kingma2013auto} and GANs~\cite{goodfellow2014generative}.

Another milestone for point cloud completion is the application of transformer-based methods. Due to their efficiency with unordered data, transformers are well-suited for point cloud processing tasks.  PCT~\cite{guo2020pct} formulates the completion task as set-to-set translation, utilizing transformers for the first time in this context. Point Transformer~\cite{pointtransformer} adopted self-attention to point clouds and suggested point transformers for various tasks. PoinTr~\cite{yu2021pointr} improved the transformer architecture for point clouds by introducing a modified attention module. The model is based on point proxies, groups of points with position embedding, and a transformer architecture to generate a complete shape.SnowflakeNet~\cite{xiang2021snowflakenet} employs an encoder-decoder architecture based on transformers to complete point clouds coarse-to-fine. Each level of the decoder architecture inputs a prediction of a subset of the complete point cloud generated by the previous level and splits the input into child points to generate a finer prediction. Similarly, Seedformer~\cite{zhou2022seedformer} introduces a new representation called patch seeds, consisting of points and their features, to be used in the decoding step. Unlike SnowflakeNet, Seedformer’s decoder levels utilize point features while generating a new set of complete point clouds. AdaPointTr~\cite{yu2023adapointr} builds on top of PoinTr by incorporating a denoising task and an adaptive query generation mechanism. Finally, AnchorFormer \cite{chen2023anchorformer} utilizes points to aid shape completion. It employs points around the shape to capture regional information about objects. These points are then used to reconstruct fine-grained objects utilizing a modulation scheme. 

Although these models predict completion well, they rely on aligning objects into canonical coordinates and fail when local coordinates are unknown, limiting their effectiveness in more generalized and unstructured scenarios. This reliance on known local coordinates and failure to predict under arbitrary rotations highlights a significant challenge that our work aims to address.

%% file: sections/03_method.tex
\section{Methodology}
\begin{figure*}
    \centering
    \includegraphics[width=\linewidth]{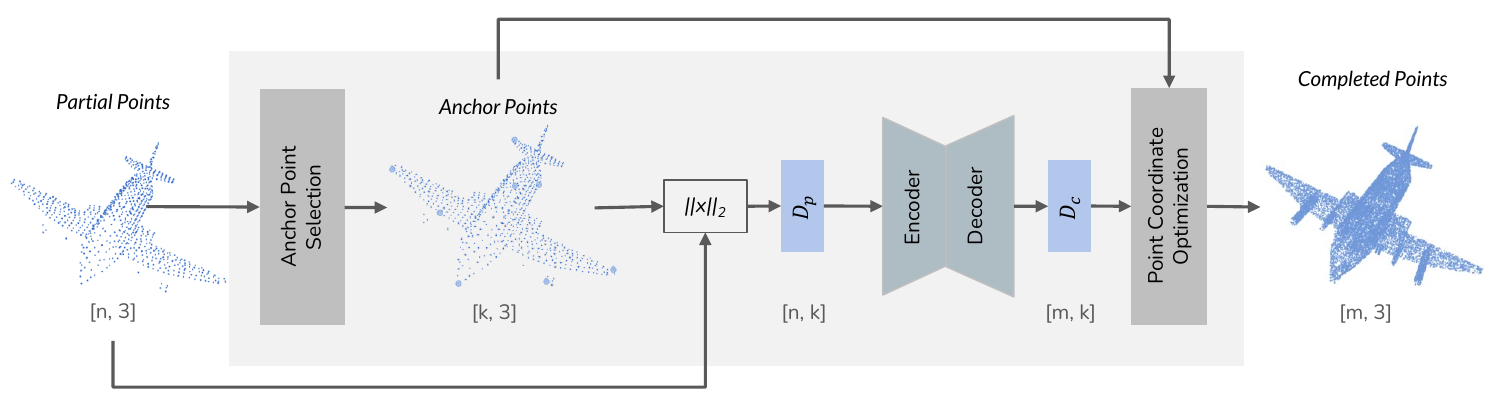}
    \caption{The overall pipeline for ESCAPE model. Initially, we extract $k$ anchor points to construct rotation-invariant features as input to a transformer-based encoder-decoder architecture. The transformer is specially modified to predict the distance between points in the complete geometry and the extracted anchor points. It simultaneously constructs complete geometry with $m$ points and predicts the distance to anchor points. Finally, an optimization procedure has been utilized to find the coordinates of the complete shape.}
    \label{fig:pipeline}
\end{figure*}
Our method addresses point cloud completion by leveraging a combination of rotation-invariant features and a specially designed transformer model. Initially, anchor points and rotation-invariant features, point distances to anchors, are extracted from the partial input point cloud. These features are then passed into the transformer and an optimization to predict the complete geometry of the object relative to the anchor points.

It is important to note that while distances are invariant to rotational changes, the final point coordinate optimization is influenced by the input anchor points, which are affected by the rotation of the input. As a result, the final predictions are also rotated, ensuring that the entire pipeline remains rotation-equivariant. This means that the output is consistently aligned with the input rotation.

The overall architecture is shown in Figure \ref{fig:pipeline}. Below, we detail the different components of our ESCAPE model.
\subsection{Anchor Points Encoding}

Given a partial point cloud \( P = \{p_1, p_2, \dots, p_n\}, p_i \in \mathbb{R}^3 \), we select a set of anchor points \( A = \{a_1, a_2, \dots, a_k\}, a_j \in \mathbb{R}^3 \). The distances between points in \( P \) and anchor points in \( A \) are computed and stored in a distance matrix \( D_p \in \mathbb{R}^{n \times k} \), where each element \( d_{ij} \) is:
\begin{equation}
d_{ij} = \|p_i - a_j\|_2, \quad \forall i \in \{1, \dots, n\}, \forall j \in \{1, \dots, k\}.
\end{equation}
This distance-based representation provides strong theoretical guarantees for shape reconstruction, as formalized in the following theorem:

\noindent\textbf{Theorem 1 (Reconstruction Uniqueness).}
Given a set of points \( P = \{p_1, \dots, p_n\} \) and anchor points \( A = \{a_1, \dots, a_k\} \), the distance matrix \( D_p \) uniquely determines \( P \) up to rigid transformation if:
\begin{enumerate}
    \item \( k \geq d+1 \) where \( d \) is the ambient dimension (3 for 3D shapes)
    \item The anchor points are in general position
    \item \( n \geq d+1 \)
\end{enumerate}

\noindent\textbf{Proof Sketch.}
The uniqueness follows from geometric constraints:
\begin{enumerate}
    \item Each point \( p_i \) is constrained by \( k \) distances to anchor points
    \item These constraints form spherical surfaces in \( \mathbb{R}^3 \)
    \item The intersection of \( k \) spheres in general position yields a unique point
\end{enumerate}

In the supplementary, we prove our distance representation maintains constant error bounds ($\mathcal{O}(1)$), unlike vector neurons' exponential error growth ($\mathcal{O}(\alpha^L)$). The distance matrix \( D_p \) serves as the input feature set for our transformer model.

The selection of anchor points is crucial. They should be well-distributed and consistent across samples within the same object category. We use furthest point sampling (FPS) initialized from the centroid to ensure equivariance, forming \( k \) clusters. Within each cluster, we compute the Laplacian \( \Delta \) of the normal vectors \( N = \{n_1, n_2, \dots, n_n\} \) for each point \( p_i \) in the cluster:
\begin{equation}
\Delta n_i = n_i - \frac{1}{|\mathcal{N}(p_i)|} \sum_{j \in \mathcal{N}(p_i)} n_j,
\end{equation}
where \( \mathcal{N}(p_i) \) denotes the neighbors of point \( p_i \).

We estimate PCA-based curvature \( \kappa_i \) at \( p_i \), based on the covariance 
matrix \( C_i \) of neighboring normals~\cite{hoppe1992surface}:
\begin{equation}
C_i = \frac{1}{|\mathcal{N}(p_i)|} \sum_{j \in \mathcal{N}(p_i)} (n_j - \bar{n}_i)(n_j - \bar{n}_i)^T,
\end{equation}
\begin{equation}
\kappa_i = \min(\text{eig}(C_i)),
\end{equation}
where \( \bar{n}_i \) is the mean normal vector of \( \mathcal{N}(p_i) \). The point with highest curvature within each cluster is selected as an anchor point, representing salient landmarks consistent across geometrically similar samples.

\subsection{Transformer Architecture}

Our point cloud transformer model is inspired by AdaPoinTr~\cite{yu2023adapointr}, but we revised it to meet the specific requirements of rotation invariance.

First, we modify the feature extraction process used to generate point proxies. In the original architecture, a DGCNN~\cite{wang2019dynamic} model is employed to extract local neighbor features through hierarchical downsampling and processing of initial features. Instead of using absolute point coordinates, we input the distances \( d_{ij} \) to the anchor points into the DGCNN, with dimensionality handled through linear layers of size \( k_{nn} \). This modification is applied consistently across the AdaPoinTr architecture, ensuring encoding with the same anchor points throughout the network. Complete implementation details of the network configurations, can be found in the supplementary material.

The use of distances also affects the training objective. The original AdaPoinTr loss function focuses on predicting the correct coordinates for noisy points perturbed during training. Since our network predicts distances, we modified the loss function to account for the noisy distances between noisy input points and noise-free anchor points. This ensures that our network can still remove noise from the inputs, even when operating in distance space. 

Similarly, in the self-attention layer, we replaced the point coordinates with the distances to anchor points to better capture the geometric relationships in the point cloud while maintaining invariance to input rotation.

These changes rely on the intuition that distances can effectively serve as encoded point coordinates. Specifically, two points will remain neighbors in Euclidean space when their distances to anchor points are used as descriptors, effectively representing the points in a "distance space." Therefore, point coordinates can be replaced by distances to anchor points without losing the geometric relationships between the points.

The final output distances are inherently unaffected by any rotations applied to the input, ensuring both consistency and rotation invariance. To reconstruct the completed point cloud with accurate 3D geometry, we solve an optimization problem to determine the final coordinates of the 3D shape from the predicted distances.

\subsection{Point Coordinate Optimization}
To finalize the point cloud completion, the predicted distances must be converted into point coordinates. This step involves finding the coordinates of points whose distances to known anchor points match the predicted values. More formally, for each point we solve the optimization problem:
\begin{equation}
  \min_{p = (x,y,z)} \sum_{j=1}^{k} \left( \|p - a_j\|_2 - \hat{d}_{ij} \right)^2,
\end{equation}
where \(a_j\) are the coordinates of the known anchor points, \(p=(x,y,z)\) are the coordinates of the point \(p_i\) to be determined, and \(\hat{d}_{ij}\) represents the predicted distance between point \(p_i\) and anchor point \(a_j\).

Following from our reconstruction uniqueness theorem, this optimization problem has a unique solution (up to reflection) when the number of anchor points \(k \geq 4\) and they are in general position. The solution uniqueness is particularly important when dealing with incomplete input shapes, as having well-distributed anchor points helps constrain the optimization even in regions with missing data. In practice, we use the Levenberg-Marquardt algorithm~\cite{levenbergMarquardt} to solve this optimization efficiently, with initialization from the centroid of anchor points to ensure consistent convergence.

The final coordinates predicted by the pipeline retain the same orientation as the partial input cloud, ensuring that the method is rotation-equivariant. Moreover, due to the model's rotation invariance, the predicted complete object coordinates remain consistent under varying transformations of the input.

%% file: sections/04_experiments.tex
\section{Experiments}

We conduct point cloud completion experiments on different datasets with different input transformations to evaluate our method's effectiveness and robustness over varying input conditions. These experiments are performed on established benchmarks: PCN\cite{yuan2018pcn} and KITTI\cite{geiger2012cvpr} as well as additional evaluation introduced on OmniObject\cite{wu2023omniobject3d} dataset to demonstrate generalizability and robustness on real-world object scans. Results demonstrate the effectiveness of our approach across many datasets, showcasing superior performance compared to existing models when presented with non-canonical inputs.

For all of our experiments, we refrained from applying rotations during training and evaluated our models by applying random rotations across three dimensions. Only in OmniObject experiments we did not rotate the input point clouds and test the methods' capability of handling arbitrary rotations originating from the projection of the depth maps with unknown extrinsic parameters.

\subsection{Training Setup}
We used the Pytorch framework and the Adam optimizer for our implementation with $\beta_1=0.9$ and $\beta_2=0.999$. We initialized the learning rate at 0.001 and utilized a learning rate scheduler that multiplied the learning rate by 0.98 in every 15 epochs. We train our models until the validation loss does not improve over the last 30 epochs and a maximum of 200 epochs. The training takes approximately 10 hours on a single NVIDIA RTX 3090 GPU. For all experiments, we use $k=8$ anchor points, input point clouds with $n=2048$ points, and generate completed shapes with $m=16384$ points.

For training, we adapt the Chamfer Distance (CD) to operate in our distance space. Given two distance 
matrices $D_1, D_2 \in \mathbb{R}^{n \times k}$ representing distances to $k$ anchor points, we define 
the Distance Matrix Chamfer Distance (DMCD) as:
\begin{equation}
\begin{split}
DMCD(D_1, D_2) = & \frac{1}{n_1} \sum_{i=1}^{n_1} \min_{j} \|\mathbf{d}^1_i - \mathbf{d}^2_j\|_1 \\ 
                  & + \frac{1}{n_2} \sum_{j=1}^{n_2} \min_{i} \|\mathbf{d}^2_j - \mathbf{d}^1_i\|_1
\end{split}
\end{equation}
where $\mathbf{d}^1_i, \mathbf{d}^2_j$ are distance vectors to anchor points. Our loss is $L = DMCD(\hat{D_c}, D_c)$, where $\hat{D_c}$ and $D_c$ are predicted and ground truth distance matrices respectively.

\subsection{The PCN Benchmark}

\textbf{The PCN Dataset} consists of 8 categories derived from the ShapeNet dataset and includes numerous instances of complete and partial point clouds. Partial clouds are derived from complete clouds by back-projecting depth images from 8 different viewpoints with varying numbers of points. Adhering to established conventions, we upsampled/downsampled 2048 points from the input to construct our inputs and generate 16,384 points as the final output.

Following the existing works, we used the PCN dataset to evaluate the models’ shape completion capabilities on rotated inputs to assess the performance degradation of the models under rotation. To generate them, we selected random degrees between 0-180 for all three axes and applied these rotations simultaneously. To evaluate existing non-equivariant methods, we applied PCA~\cite{abdi2010principal} to align the rotated point clouds, thereby normalizing the inputs before completion. After predicting the complete geometry, we applied the inverse rotation to align the completed shapes with the ground truth. In supplementary material we also integrate ConDor following \cite{sen2023scarp} to canonicalize point clouds before completion. 

\begin{figure*}
    \centering
    \includegraphics[width=0.93\textwidth]{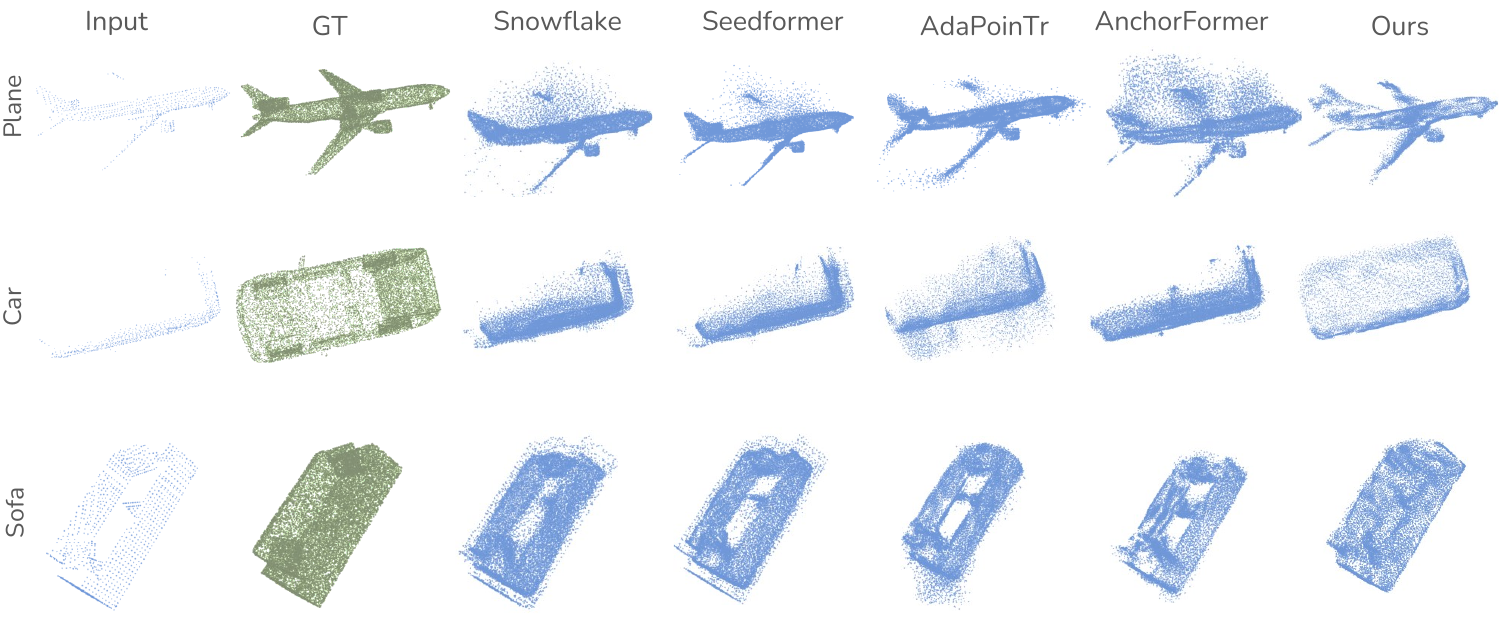}
    \caption{Qualitative comparison of models trained on PCN dataset and tested with rotated inputs. Each row contains the input to the model in its first column. Every even row contains the rotated input of the preceding row.}
    \label{fig:vis_res_all}
\end{figure*}

\textbf{Evaluation. } We used Chamfer Distance under the L1-norm as the evaluation criterion for point cloud completion. It measures the distance between two unordered sets and is commonly used as a metric for the PCN dataset benchmark. Following existing methods, we used the same train/validation/test splits of the PCN dataset and reported the value of the Chamfer Distance with L1 norm, multiplied by 1000. Table: \ref{table:completion_res} shows the detailed results, and Figure \ref{fig:vis_res_all} depicts qualitative comparison with other methods. 

Our method is the only model where prediction is unaffected by the input rotation and achieves rotation-equivariance. As shown in Figure \ref{fig:vis_res_all}, ESCAPE can achieve high-resolution outputs for planes and cars that fall under shape categories without variation.

Other existing methods are bound to canonical shape representation and therefore face significant performance degradation in completion. Their predictions are subject to high noise and structural deformations. As shown in Figure \ref{fig:vis_res_all}, most predictions become barely recognizable. Interestingly, models with better reported completion results (AdaPoinTr, AnchorFormer) achieve worst results under this settings. This phenomenon is that existing methods overfit the dataset they trained on and, therefore, lack robustness against real-world scenarios. 
On the other hand, ESCAPE can yield equivariant predictions under rotation, superior to all other methods. This makes ESCAPE applicable to scenarios that require non-canonical point processing. 

\begin{table}[ht]
\caption{Results on PCN dataset. We use CD-L1($\times1000$) as an evaluation metric and report the results for rotated inputs. The best results are written with bold letters}
\resizebox{\linewidth}{!}{
\begin{tabular}{lllllll}
 & Snowflake\cite{xiang2021snowflakenet} & Seedformer\cite{zhou2022seedformer} & PointTr\cite{yu2021pointr} & AdaPoinTr\cite{yu2023adapointr} & AnchorFormer\cite{chen2023anchorformer} & Ours \\ \hline
Plane         & 72.71  & 76.19  & 13.03  & 12.10 & 11.88 & \textbf{8.6} \\
Cabin        &  85.81 &  85.99 &47.97 & 50.35  & 32.93 & \textbf{13.62} \\
Car          &  78.76 &  82.28 &  37.42  & 40.90 &  28.97 &   \textbf{10.43} \\
Chair        &  64.57 &  66.28 &  30.53  & 37.24 &  34.94 &   \textbf{10.71} \\
Lamp         &  141.04 &  148.67 &  19.01 & 19.77  &  17.73 & \textbf{8.14} \\
Sofa         & 73.63 &  77.47 &  44.13  & 49.46 &  33.89 &  \textbf{13.86} \\
Table        &  87.31 &  89.58 &  29.03  & 37.74 & 36.57 &   \textbf{9.23} \\
Boat         &  106.97 &  110.72 &  20.51  & 20.60 &  16.26 & \textbf{10.00} \\ \hline
Avg          &  88.85 &  92.15 &  30.20  & 33.52 &  26.65 &    \textbf{10.58} \\ \hline
\end{tabular}
}
\label{table:completion_res}
\end{table}

\subsection{OmniObject3D Benchmark}

Our next experiment focuses on a real-world scenario that requires reconstructing real-world objects from depth maps whose pose parameters are unknown. To create the experiment, we collected seven categories from the OmniObject~\cite{wu2023omniobject3d} dataset that are also included in the PCN~\cite{yuan2018pcn} dataset. Each category has multiple samples with a complete point cloud and 100 depth maps from different viewpoints. We back-projected these depth maps to obtain partial point clouds with unknown orientations. Before feeding the models with the inputs, we applied normalization on partial point clouds to match their distribution with the PCN dataset. We used the inverse of the input transformation to the complete point clouds to evaluate the completion performance.

We observed that the OmniObject dataset is challenging for point cloud completion for two reasons. First, it contains objects with varying geometry dissimilar to their PCN pairs. Secondly, incorrect depth and intrinsic parameters for some samples lead to isolated points in the input point clouds, significantly affecting the method's completion ability. We observed that the mentioned challenges caused the metric value to degrade, misleading the models' performances. Therefore, to remove these outliers, we report the median per category instead of the average of each sample in each category.  

\begin{table}[ht]
\caption{Results on OmniObject dataset. We report median CD-L1($\times1000$) per category as an evaluation metric and report the average of all categories. The best results for both input types is written with bold letters}
\resizebox{\linewidth}{!}{
\begin{tabular}{lllllll}
\hline
      & Snowflake\cite{xiang2021snowflakenet} & Seedformer\cite{zhou2022seedformer} & PoinTr\cite{yu2021pointr} & AdaPoinTr\cite{yu2023adapointr} & AnchorFormer\cite{chen2023anchorformer} & Ours \\ \hline
Plane & 16.36     & \textbf{13.90}      & 15.49  & 17.53     & 18.20        &  16.00    \\
Cabin & 37.45     & 34.61      & 46.13  & 54.02     & 51.00        & \textbf{12.09}     \\
Car   & 34.38     & 32.70      & 40.07  & 49.97     & 47.69        &   \textbf{26.3}   \\
Chair & 23.12     & 21.08      & 23.12  & 31.74     & 27.84        &  \textbf{7.2}     \\
Lamp  & 54.78     & 59.02      & 54.78 & 67.36     & 72.13       &  \textbf{49.85}    \\
Sofa  & 28.36     & 26.16      & 28.36  & 38.88     & 38.42        &  \textbf{9.7}    \\
Boat  & 21.71     & 19.47      & 22.88  & 22.49     & 25.60        &  \textbf{10.62}    \\ \hline
Avg   & 31.14     & 29.56      & 32.97  & 39.57     & 40.12        &  \textbf{18.82} \\ \hline
\end{tabular}
}

\label{table:omniRes}
\end{table}

\begin{figure}
    \centering
    \includegraphics[width=\linewidth]{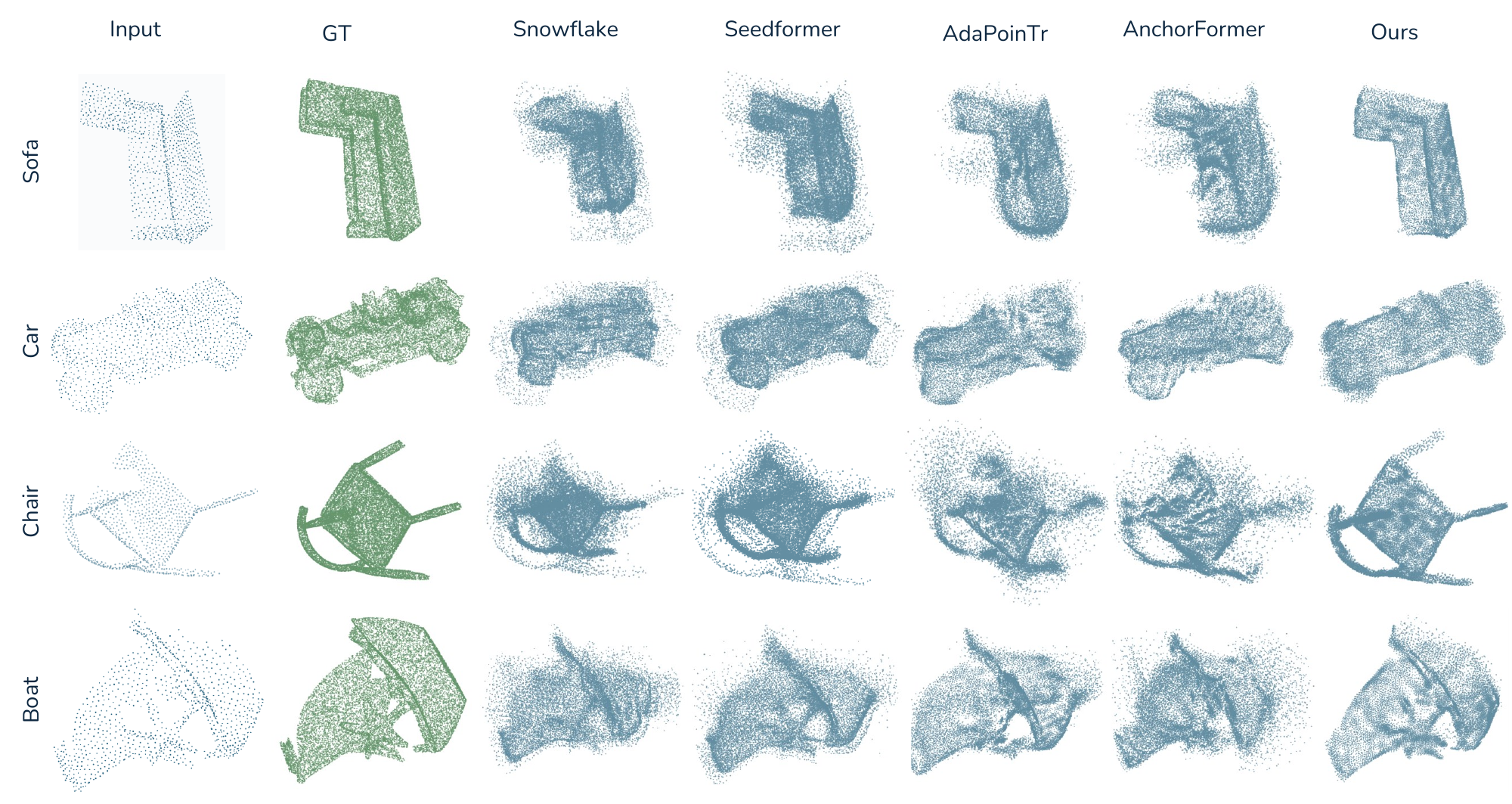}
    \caption{Qualitative comparison of models trained on PCN dataset and tested on OmniObject dataset. Each row contains the input to the model in its first column. }
    \label{fig:vis_omni}
\end{figure}

Similar to the PCN dataset, our method is the only model unaffected by the input rotation. Results in Table \ref{table:omniRes} show that it is capable of handling unknown object poses and is still able to reconstruct objects in high resolution. Qualitative results in Figure \ref{fig:vis_omni} are evidence of our method's capability of handling arbitrary poses while existing methods failed to generate a structured geometry. This further proves that are model does not require prior pose estimation to align the object frame and can directly complete the occluded regions. We refer readers to supplementary material for qualitative results.

\subsection{KITTI Cars Benchmark}

Another real-world application includes predicting complete car object geometries in the KITTI dataset, which contains incomplete point clouds from LiDAR scans in real-world scenes. We pretrain our model only with car images from the PCN dataset. As the complete point clouds are not available in this dataset, we reported the MMD and Fidelity score for inputs with certain rotation: a random rotation around a single axis to mimic the movement of a vehicle. Table \ref{tab:resKitti} shows the performance of the models. 

The MMID results in the rotated samples show that existing methods fail to complete the cars when a simple yet realistic rotation is applied to the inputs. 

\begin{table}[ht]
\caption{Results on KITTI Cars dataset for rotated inputs. Fidelity and MMID metrics are calculated using CD-L2 (x1000) distance.}
\centering
\begin{tabular}{lllll}
\hline
          & Fidelity & MMID     \\ \hline
PoinTr\cite{yu2021pointr}  & \textbf{0.0}  & 6.15          \\
Snowflake\cite{xiang2021snowflakenet} & 1.77  & 16.08         \\
Ours      & 1.81         & \textbf{5.93} \\ \hline
\end{tabular}
\label{tab:resKitti}
\end{table}


\subsection{Ablation Studies}

\begin{table}[t]

\caption{Comparison of different rotation equivariant methods. For SCARP, we evaluated the performance only using the common categories.}
\resizebox{\linewidth}{!}{
\label{tab:rotation_methods}  
\centering
\begin{tabular}{lllllllll}
\hline
Method & Plane        & Cabin         & Car           & Chair         & Lamp         & Sofa          & Boat         & Avg           \\ \hline
SCARP\cite{sen2023scarp}   & 104.4        & -         & 135.9         & 147.1         & -        & -         & 108.4        & 124.0         \\
Snowflake (VN)\cite{deng2021vector}   & 10.65        & 26.64         & 11.92         & 17.86         & 22.82        & 22.90         & 18.19        & 18.62         \\
Snowflake (PPF)  & 8.68        & 19.68         & 10.95         & 19.36         & 25.96    & 18.62 & 16.07 & 17.46         \\
Ours   & \textbf{8.6} & \textbf{13.62} & \textbf{10.43} & \textbf{10.71} & \textbf{8.14} & \textbf{13.86} & \textbf{10.0} & \textbf{10.58} \\ \hline
\end{tabular}
}
\end{table}

\subsubsection{Other Encodings}
\begin{figure}
    \centering
    \includegraphics[width=0.9\linewidth]{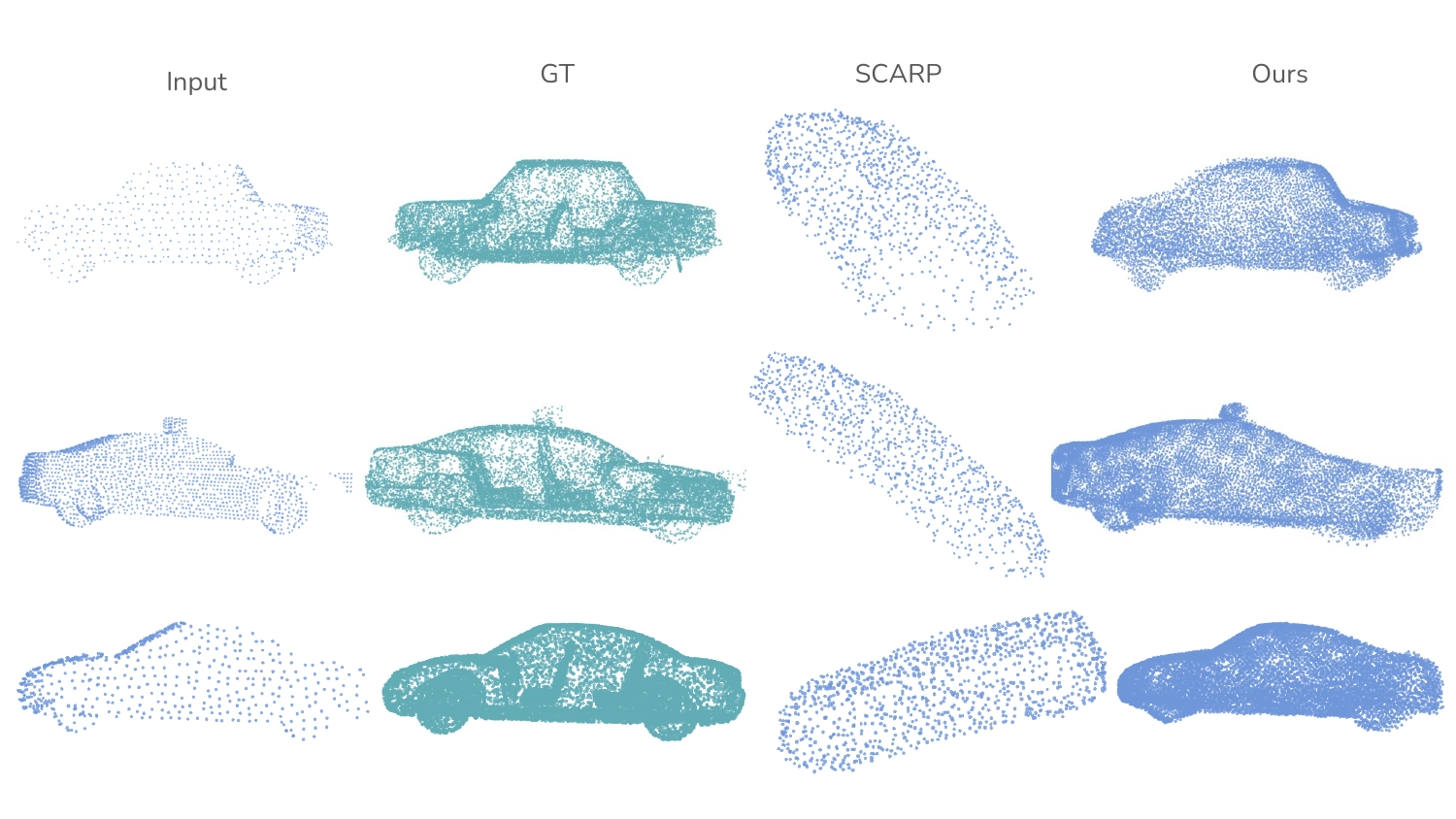}
    \caption{Qualitative comparison of SCARP and ESCAPE on PCN dataset. }
    \label{fig:comp_scarp}
\end{figure}

Shape completion literature lacks robust methods against input transformations to be a baseline for our process. To fill this gap and provide rotation equivariance encodings as a baseline to our method, we proposed three methods: (i) We experimented with SCARP~\cite{sen2023scarp} on the common categories (ii) We modified the DGCNN architecture built with Vector Neuron layers for point cloud completion by combining its rotation-invariant encoder with the decoder of the Snowflake. Referred as Vector Neurons in Table~\ref{tab:rotation_methods}. (iii) Snowflake~\cite{xiang2021snowflakenet} network processing point pair features(PPFs) and modified to be rotation-invariant. Referred as PPF-Snowflake in Table \ref{tab:rotation_methods}.
Results in Table \ref{tab:rotation_methods} demonstrate that our method consistently outperforms the baselines across all categories, highlighting its robust and richer equivariant encoding. ESCAPE significantly surpasses SCARP in every category. SCARP struggles to capture finer shape details, often predicting only a coarse object geometry as visualized in Figure \ref{fig:comp_scarp}. Additionally, SCARP fails to estimate the rotation of the partial cloud accurately, even for inputs that are not rotated, resulting in a significant decline in its performance.

\subsubsection{Anchor Point Selection}

As part of our ablation study, we further investigate our anchor point selection algorithm and analyze the characteristics of optimal point sets. We evaluated the performance of different anchor point selection methods on a subset of the PCN dataset. The results are presented in Table \ref{tab:anchorAblation}. For visual reference, we refer the reader to the supplementary material, which illustrates the points selected by the algorithms.

\begin{table}[H]
\caption{Results of different anchor point selection algorithms on a subset of the PCN dataset. CD-L1 (x1000) distance is reported.}
\centering
\begin{tabular}{lll}
\hline
\textbf{Algorithm} & \textbf{Threshold/Radius} & \textbf{CD-L1 Score} \\ \hline
\textit{Clustering}   & 0.0       & 14.80    \\
\textit{Clustering}   & 0.5       & 15.73    \\ \hline
\textit{Ball Query}   & 0.05      & 14.23    \\
\textit{Ball Query}   & 0.075     & \textbf{13.58}    \\
\textit{Ball Query}   & 0.1       & 15.00    \\
\textit{Ball Query}   & 0.15      & 15.76    \\ \hline
\textit{FPS}          & -         & 14.74   \\ \hline
\end{tabular}
\label{tab:anchorAblation}
\end{table}

In these experiments, we analyse the effect of anchor point selection based on the curvature values of points. First, we use the FPS algorithm to select $a$ anchor points, then use them as cluster centers, assigning all other points to their closest cluster. We choose the point with the maximum curvature as the anchor point for each cluster, provided its curvature exceeds a predefined threshold. The results of this algorithm, referred to as "Clustering," are shown in Table \ref{tab:anchorAblation}. 

For comparison, the final row of Table \ref{tab:anchorAblation}, labeled "FPS," shows the results of using the FPS algorithm alone without further refinement. Both results demonstrate the importance of well-distributed anchor points around the shape for the model’s success. Selecting anchor points from clusters allows points to move away from their initial positions, thereby covering less area of the object’s geometry.

Additionally, we limit the refinement of anchor points to a maximum distance (radius) from their initial positions. The rows labeled "Ball Query" in Table \ref{tab:anchorAblation} show the results for different radius values. The results indicate that a larger radius does not improve performance, as it allows too much flexibility for anchor points to shift around the shape, resulting in less coverage of critical regions. This observation aligns with the conclusions from previous experiments.

The "Ball Query" results further show that selecting points with respect to their curvature is beneficial. By allowing points to move a small distance, such as 0.075, from their initial positions, we could choose better anchor points, resulting in improved point cloud completion performance.


%% file: sections/05_conclusion.tex
\section{Conclusion}

In this work, we presented Equivariant Shape Completion via Anchor Point Encoding (ESCAPE), a novel method for achieving rotation-equivariant shape completion. Our approach tackles the key challenge of reconstructing object geometries from various orientations by leveraging a distance-based feature encoding inspired by the D2 shape distribution. 

Through comprehensive experiments on the PCN, KITTI, and OmniObject datasets, ESCAPE consistently outperformed existing models, mainly when dealing with rotated input data. This demonstrates the method's robustness in scenarios where input orientation is not controlled or known in advance. ESCAPE provides a practical solution for dynamic environments, such as robotic manipulation and real-time object recognition, by enabling effective shape completion without requiring prior knowledge of object orientation or additional pose estimation modules. The combination of rotation-equivariant anchor selection and a transformer-based architecture enables ESCAPE to reconstruct objects with high precision, capturing fine details and maintaining geometric consistency throughout the process.

%% file: sections/supplementary.tex
\appendix
\section{Robustness Analysis}

\subsection{Error Bounds}
This section provides detailed proofs for the theoretical guarantees mentioned in the main paper regarding our distance-based representation's error bounds compared to other equivariant features such as Vector Neurons.

In Vector Neurons, each layer performs transformation:
\begin{equation}
    \mathbf{v}_{\text{out}} = \mathbf{R}(\mathbf{W}\mathbf{v}_{\text{in}})
\end{equation}
where $\mathbf{R}$ is a rotation matrix and $\mathbf{W}$ are learned weights. For an input perturbation $\boldsymbol{\epsilon}$, the error after $L$ layers propagates as:
\begin{equation}
    \|\mathbf{e}_L\| \leq \prod_{i=1}^L \|\mathbf{W}_i\| \cdot \|\mathbf{R}_i\| \cdot \|\boldsymbol{\epsilon}\| = \mathcal{O}(\alpha^L)
\end{equation}
where $\alpha = \max_i \|\mathbf{W}_i\| \cdot \|\mathbf{R}_i\| > 1$ typically.

For our distance-based representation, given points $P$ and anchor points $A$, a perturbation $\boldsymbol{\epsilon}$ affects distances as:
\begin{equation}
    |d(p + \boldsymbol{\epsilon}, a) - d(p, a)| \leq \|\boldsymbol{\epsilon}\|
\end{equation}
by the triangle inequality. This leads to our key result:

\begin{theorem}[Error Bounds]
For input perturbation $\boldsymbol{\epsilon}$, our distance matrix representation maintains constant error bounds:
\begin{equation}
    \|D(P + \boldsymbol{\epsilon}, A) - D(P, A)\|_\infty \leq \|\boldsymbol{\epsilon}\|_2
\end{equation}
independent of network depth.
\end{theorem}

This theoretical advantage is empirically validated in Section 4 of the main paper, where we demonstrate superior robustness on real-world datasets that are based on VectorNeurons.

\subsection{Performance under Noise}
To validate our theoretical error bounds empirically, we evaluate ESCAPE's robustness by adding Gaussian noise with varying standard deviations $\sigma$ to input point clouds. As shown in Table \ref{table:noiseRes}, our method maintains stable performance under increasing noise levels ($\sigma$  = 0.001 to 0.004), with average CD-L1 scores only degrading from 10.58 to 10.99. 

\begin{table}[H]
\caption{Results of ESCAPE under input noise.}
\centering
\begin{tabular}{llllll}
\hline
Category & $\sigma$ & 0.0      & 0.001    & 0.002    & 0.004   \\ \hline
Plane    & & \textbf{8.60}     & 8.66     & 8.83     & 8.71    \\
Cabin    & & 13.62    & 13.74    & \textbf{13.45}    & 13.67   \\
Car      & & \textbf{10.43}    & 10.46    & 10.47    & 10.76   \\
Chair    & & 10.71    & 10.72    & \textbf{10.68}    & 11.08   \\
Lamp     & & 8.14     & \textbf{8.11}     & 8.30     & 8.87    \\
Sofa     & & \textbf{13.86}    & 14.00    & 14.05    & 14.73   \\
Table    & & \textbf{9.23}     & 9.28     & 9.40     & 9.48    \\
Boat     & & \textbf{10.00}    & 10.01    & 10.01    & 10.55   \\ \hline
Avg      & & \textbf{10.58}    & 10.62    & 10.65    & 10.99   \\ \hline
\end{tabular}
\label{table:noiseRes}
\end{table}

\subsection{Different Level of Partiality}

\begin{figure}[h]
    \centering
    \includegraphics[width=\linewidth]{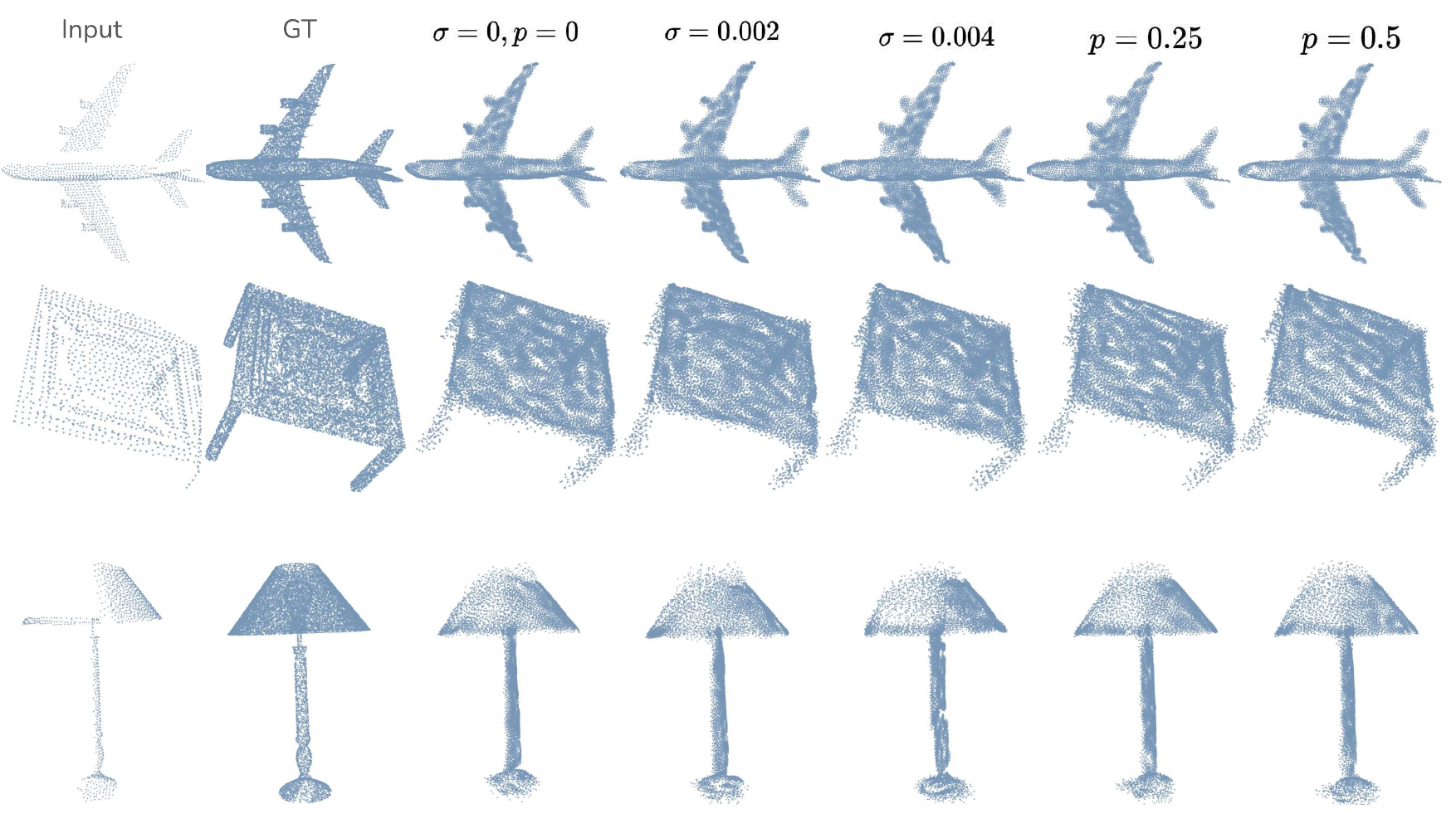}
    \caption{Quantitative results of additive noise and point removal experiments.}
    \label{fig:anchorRobustness}
\end{figure}

Similar to the noise experiment, here we exclude a random portion of the input points, determined by a removal ratio of $p$. The results of point removal are shown in Table \ref{table:removeRes}. The experiment demonstrate that our method is robust to input perturbations, maintaining consistent performance across varying conditions.
The robustness of ESCAPE is further illustrated in Figure \ref{fig:anchorRobustness}.  ESCAPE reliably handles input noise and maintains output quality, even when large portions of input points are removed.

\begin{table}[H]
\caption{Results of removing portion of the input points.}
\centering
\begin{tabular}{llllll}
\hline
Category& $p$ & 0.0    & 0.1   & 0.25  & 0.5   \\ \hline
Plane   & & 8.60   & 8.49  & \textbf{8.32}  & 8.92  \\
Cabin   & & 13.62  & 12.69 & \textbf{12.75} & 13.00 \\
Car     & & 10.43  & \textbf{10.39} & 10.81 & 10.41 \\
Chair   & & \textbf{10.71}  & 11.43 & 11.21 & 11.36 \\
Lamp    & & \textbf{8.14}   & 8.72  & 8.70  & 8.84  \\
Sofa    & & 13.86  & \textbf{13.74} & 13.80 & 13.97 \\
Table   & & \textbf{9.23}   & 9.96  & 10.06 & 10.65 \\
Boat    & & 10.00  & 9.96  & \textbf{9.83}  & 10.02 \\ \hline
Avg     & & \textbf{10.58}  & 10.68 & 10.69 & 10.90 \\ \hline
\end{tabular}
\label{table:removeRes}
\end{table}

\subsection{Visualization of KITTI Dataset}
\begin{figure}[h]
    \centering
    \includegraphics[width=\linewidth]{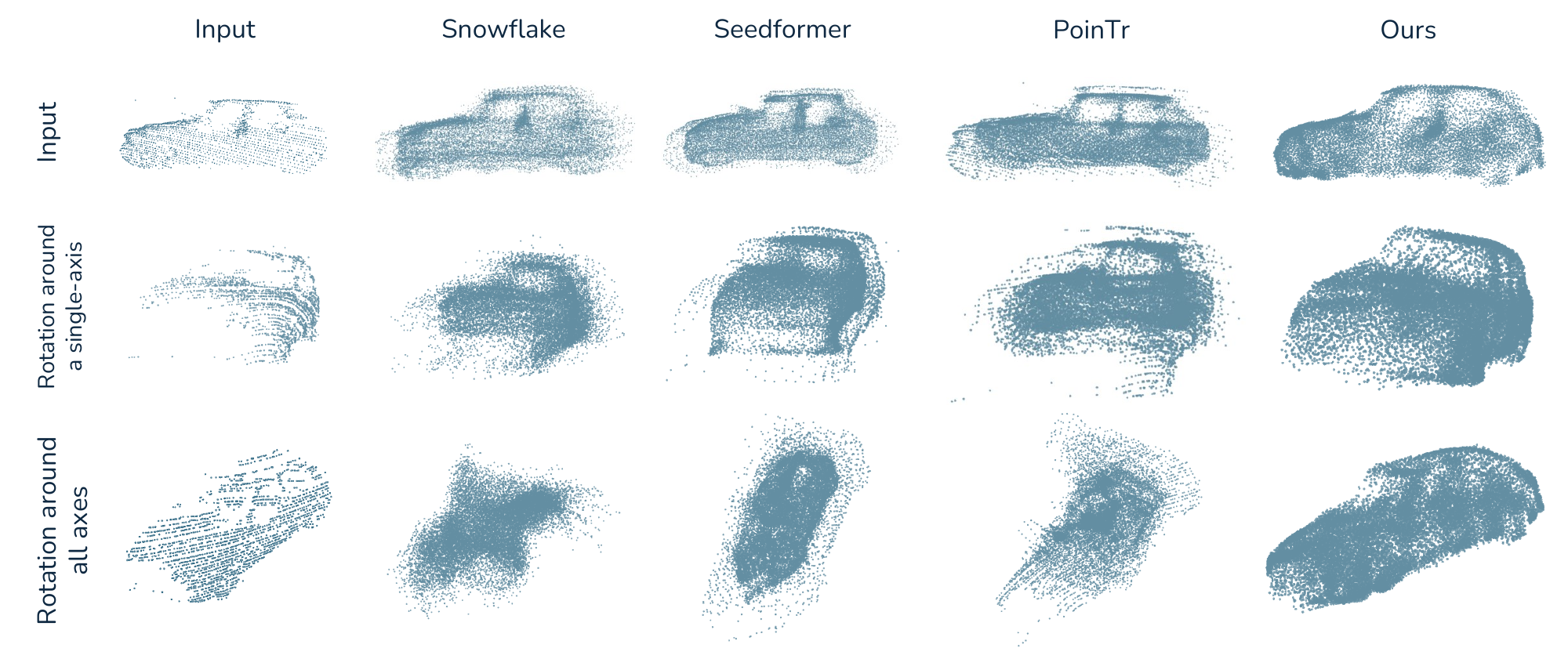}
    \caption{Qualitative comparison of models fine-tuned on PCN dataset cars category and tested on KITTI dataset. The first row contains the original input. The second row contains a single-axis rotation of the input mimicking the movement of a car. The final row contains the partial input rotated in all three axes. }
    \label{fig:vis_kitti}
\end{figure}

To further demonstrate robustness on real-world data, we evaluate on the KITTI dataset using models pretrained on PCN car shapes. Unlike controlled environments, KITTI presents noisy, partial scans from actual LiDAR sensors with unknown object orientations. As shown in Figure \ref{fig:vis_kitti}, ESCAPE maintains consistent completion quality under various orientations, including single-axis rotations mimicking vehicle movement and arbitrary three-axis rotations. This validates our method's practical applicability in real-world scenarios where clean, canonically-oriented inputs cannot be guaranteed.

\section{Anchor Point Analysis}
\subsection{Deterministic FPS}

In our method, we modified farthest point selection algorithm to make it deterministic and obtain exactly same results for the same input independent of rotation. To remove the randomness of FPS, we changed the selection of the first point by finding the point most distant from the center point calculated as $C = \frac{1}{|P|} \sum_{p_i \in P} {p_i}$. Selection of all other points executed normally. With this modification, our method's performance stays constant even under input rotation.

\subsection{Comparison with Learned Keypoints}

ESCAPE's modular design supports integration with any set of keypoints generated by various algorithms. For comparison, we selected two self-supervised keypoint detection methods: SNAKE~\cite{zhong2022snake} and SkeletonMerger~\cite{shi2021skeleton} and benchmarked them against our curvature-based keypoint selection approach. We trained class-specific models to generate keypoints using SkeletonMerger and ensured the number of keypoints matched those used in ESCAPE. Similarly, we utilized its pre-trained weights for SNAKE and applied Farthest Point Sampling (FPS) to select a subset of predicted keypoints, aligning their count with ESCAPE's. The results are given in Table \ref{table:diffKeypointTypesRes}. 
The results demonstrated that our anchor points outperformed learning-based counterparts, achieving superior performance on rotated and canonical input datasets.

\begin{table}[H]
\caption{Results on PCN dataset with different set of keypoints. We use CD-L1($\times1000$) as an evaluation metric and report the results for normal and rotated inputs. Best results for both input type is written with bold letters.}
\resizebox{\linewidth}{!}{
\begin{tabular}{lllll}
\hline
Category & Keypoints & Skeleton~\cite{shi2021skeleton}       & SNAKE~\cite{zhong2022snake}          & Anchors        \\ \hline
Airplane & canonical    & \textbf{6.73}  & 10.47          & 8.6            \\
         & rotated   & 19.17          & 10.66          & \textbf{8.6}   \\
Cabin    & canonical    & 28.14          & 13.86          & \textbf{13.62} \\
         & rotated   & 30.23          & 13.64          & \textbf{13.62} \\
Car      & canonical    & \textbf{9.91}  & 10.35          & 10.43          \\
         & rotated   & 20.89          & 10.49          & 10.43          \\
Chair    & canonical    & \textbf{9.98}  & 10.57          & 10.71          \\
         & rotated   & 19.64          & \textbf{10.80} & 10.71          \\
Lamp     & canonical    & \textbf{8.02}  & 8.67           & 8.14           \\
         & rotated   & 11.51          & 9.08           & \textbf{8.14}  \\
Sofa     & canonical    & \textbf{12.74} & 14.65          & 13.86          \\
         & rotated   & 26.15          & 14.79          & \textbf{13.86} \\
Table    & canonical    & \textbf{8.19}  & 9.54           & 9.23           \\
         & rotated   & 21.24          & 9.38           & \textbf{9.23}  \\
Boat     & canonical    & \textbf{9.59}           & 10.97          & 10.00 \\
         & rotated   & 17.66          & 11.25          & \textbf{10.00} \\ \hline
Avg  & canonical    & 11.67          & 11.14          & \textbf{10.58} \\
         & rotated   & 20.82          & 11.27          & \textbf{10.58} \\ \hline
\end{tabular}
\label{table:diffKeypointTypesRes}
}
\end{table}

\subsection{Effect of Number of Anchor Points}
We analyze how the number of anchor points affects reconstruction quality during optimization. In this experiment, we compare using different subsets (k = 3, 4, and 6) of anchor points against our full set of 8 points, while maintaining the predicted distances for each subset. The results in Table \ref{table:anchorPointNum} demonstrate a clear trend: increasing the number of anchor points improves reconstruction quality. With fewer anchor points, the optimization becomes more sensitive to noise and prediction errors, as there are fewer geometric constraints. However, using too many anchor points increases model complexity and computational overhead. Through extensive experiments, we found that k = 8 provides the optimal balance between reconstruction accuracy and computational efficiency.

\begin{table}[ht]
\caption{Results of having different number of anchor points in the optimization.}
\begin{tabular}{llllll}
\hline
Category & K & 3 & 4     & 6     & 8              \\ \hline
Airplane &   & 10.21  & 9.98  & 8.68  & \textbf{8.6}   \\
Cabin    &   & 14.87  & 14.77 & 13.54 & \textbf{13.62} \\
Car      &   & 12.54  & 12.50 & 10.84 & \textbf{10.43} \\
Chair    &   & 13.13  & 13.37 & 11.10 & \textbf{10.71} \\
Lamp     &   & 9.70  & 9.79  & 8.43  & \textbf{8.14}  \\
Sofa     &   & 16.19  & 16.32 & 14.28 & \textbf{13.86} \\
Table    &   & 12.78  & 12.65 & 9.51  & \textbf{9.23}  \\
Boat     &   & 11.18  & 11.13 & 9.96  & \textbf{10.00} \\ \hline
Avg      &   & 12.58  & 12.57 & 10.79 & \textbf{10.58} \\ \hline
\end{tabular}
\label{table:anchorPointNum}
\end{table}

\section{Extended Comparisons}
\subsection{Performance on Canonical Inputs}
While our method prioritizes rotation equivariance through invariant features instead of point coordinates, this design choice incurs a small performance trade-off on canonical inputs. As shown in Table \ref{table:canonicalCompletion_res}, existing methods achieve slightly better CD-L1 scores on canonical inputs by exploiting aligned data, but their performance degrades significantly (up to 14x worse) under rotation. Notably, methods with better canonical performance show more severe degradation under rotation, suggesting dataset memorization rather than true geometric understanding. This observation is further validated by OmniObject results, where models with lower PCN scores demonstrated better generalization to real-world data. In contrast, ESCAPE maintains consistent performance (10.58) regardless of orientation, demonstrating the value of our rotation-invariant representation.

\begin{table*}[ht]
\caption{Results on PCN dataset. We use CD-L1($\times1000$) as an evaluation metric and report the results for canonical inputs versus rotated frame. The best results for both input types are written with bold letters.}
\resizebox{\textwidth}{!}{
\begin{tabular}{lllllll}
\hline
      & \multicolumn{1}{c}{Snowflake~\cite{xiang2021snowflakenet}} & \multicolumn{1}{c}{Seedformer~\cite{zhou2022seedformer}} & \multicolumn{1}{c}{PointTr~\cite{yu2021pointr}} & \multicolumn{1}{c}{AdaPoinTr~\cite{yu2023adapointr}} & \multicolumn{1}{c}{AnchorFormer~\cite{chen2023anchorformer}} & \multicolumn{1}{c}{Ours} \\ \hline
Canonical   & 7.21                                                      & 6.74                                                    & 8.38                                           & \textbf{6.53}                                       & 6.59                                                        & 10.58                    \\ 
Rotated          &  88.85 &  92.15 &  30.20  & 33.52 &  26.65 &    \textbf{10.58} \\ \hline
\end{tabular}
}
\label{table:canonicalCompletion_res}
\end{table*}

\subsection{Point Cloud Canonicalization}
To provide a comprehensive baseline comparison, we evaluate existing methods with a learned canonicalization approach. Specifically, we employ ConDor~\cite{sajnani2022condor}, a state-of-the-art point cloud canonicalization method previously used in SCARP~\cite{sen2023scarp}, to align input shapes to a canonical frame before processing. Table \ref{table:condorResults} shows the completion results of non-equivariant methods using ConDor's alignment. Notably, while SeedFormer~\cite{zhou2022seedformer} maintains reasonable performance (14.07 CD-L1), other methods show significant degradation compared to their performance on naturally canonical inputs, with average performance drops of up to 9x. These results highlight the challenges of relying on learned canonicalization for rotation-invariant shape completion.

\begin{table*}[ht]
\caption{Results with point cloud completion models tested with canonicalized inputs using ConDor~\cite{sajnani2022condor}.}
\resizebox{\textwidth}{!}{
\begin{tabular}{llllll}
\hline
Category & Snowflake~\cite{xiang2021snowflakenet} & Seedformer~\cite{zhou2022seedformer} & PoinTr~\cite{yu2021pointr} & AdaPoinTr~\cite{yu2023adapointr} & AnchorFormer~\cite{chen2023anchorformer} \\ \hline
Plane    & 70.49     &\textbf{16.39}      & 63.81  & 65.34     & 66.15         \\
Cabin    & 57.42     & \textbf{11.72}      & 58.04  & 65.11     & 67.59         \\
Car      & 56.23     & \textbf{11.40}      & 54.14 & 56.17     & 60.81         \\
Chair    & 59.31     & \textbf{13.02}      & 60.07  & 63.73     & 67.88         \\
Lamp     & 69.61     & \textbf{17.68}      & 64.74  & 68.33     & 68.39         \\
Sofa     & 63.22     & \textbf{15.35}      & 61.70  & 66.69     & 70.06         \\
Table    & 63.35     & \textbf{14.79}      & 62.94  & 72.52     & 76.64         \\
Boat     & 58.40     & \textbf{12.21}      & 49.07  & 50.42     & 53.68         \\ \hline
Avg      & 62.26     & \textbf{14.07}      & 59.31  & 63.54     & 66.40         \\ \hline
\end{tabular}
}
\label{table:condorResults}
\end{table*}

\section{Computational efficiency}

We measured the average inference time of ESCAPE and compared to against existing methods, as summarized in Table \ref{table:inferenceTime}. The results show that our method ranks second fastest regarding the milliseconds required to predict a single sample. 

\begin{table}[ht]
\centering
\caption{Inference time of a single input with existing methods.}
\begin{tabular}{ll}
\hline
Models               & Elapsed Time \\ \hline
Snowflake\cite{xiang2021snowflakenet}            & 16.7 ms      \\
Seedformer\cite{zhou2022seedformer}           & 45.1 ms      \\
PoinTr\cite{yu2021pointr}               & 38.5 ms      \\
AdaPoinTr\cite{yu2023adapointr}            & 35.6 ms      \\
AnchorFormer\cite{chen2023anchorformer}         & 21.8 ms      \\
ESCAPE               & 19.3 ms      \\ \hline
\end{tabular}
\label{table:inferenceTime}
\end{table}

\section{Analysis of Point Selection and Distribution}
\subsection{Visualization of point curvatures}
\begin{figure}[h]
    \centering
    \includegraphics[width=\linewidth]{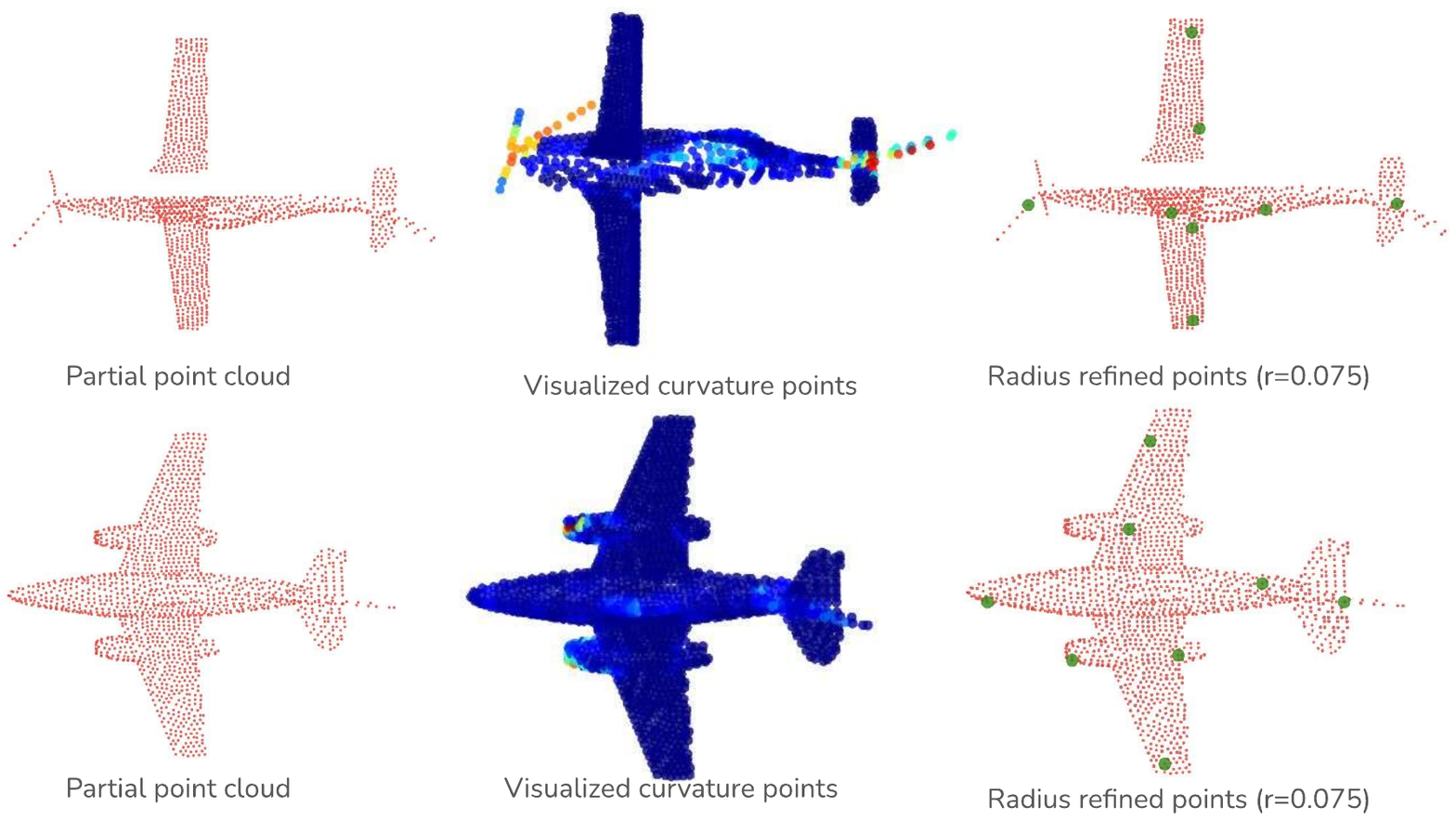}
    \caption{Heatmap of the point curvature values and derived anchor points.}
    \label{fig:curvatureVis}
\end{figure}

Figure \ref{fig:curvatureVis} illustrates the calculated curvature value for each point in the input and derived refined anchor points using the described algorithm in the main paper. The figure depicts that our algorithm can select points with high curvature and still cover all regions of the shape, hence good anchor points for point cloud completion.

\begin{figure*}[h!]
    \centering
    \includegraphics[width=\textwidth]{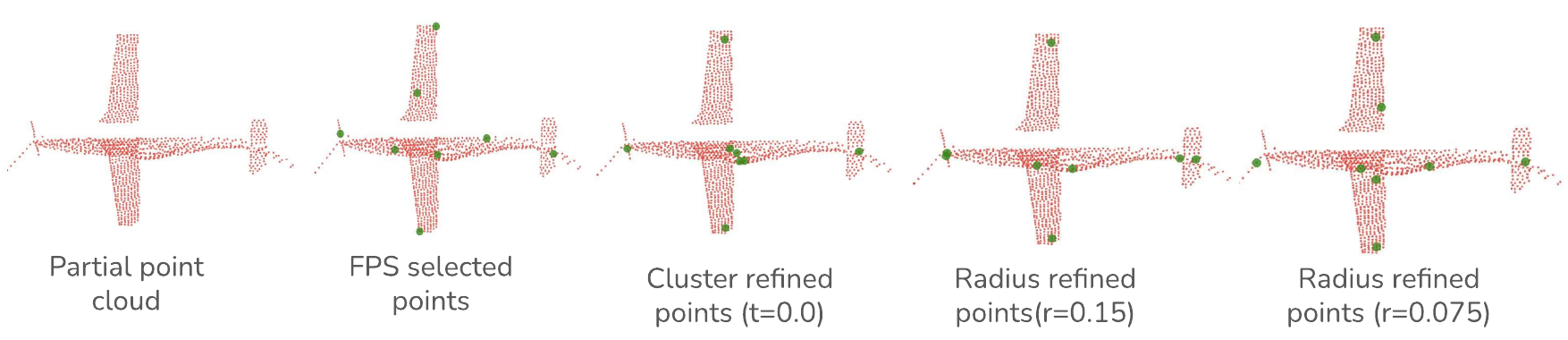}
    \caption{Comparison of different anchor point selection algorithms.}
    \label{fig:anchorComp}
\end{figure*}

\subsection{Analysis of Anchor Point Selection Strategies}
Figure \ref{fig:anchorComp} provides a comprehensive comparison of different anchor point selection strategies. Our analysis reveals that initializing with Farthest Point Sampling (FPS) provides effective coverage by selecting well-distributed points across the shape. While these points can be refined using curvature information (as shown in Figure \ref{fig:curvatureVis}), we find that the refinement process must be constrained. Without proper limits, the refinement can cause anchor points to cluster (visible in top-right and bottom-left of Figure \ref{fig:anchorComp}), reducing spatial coverage and potentially leading to fewer effective anchor points. Our proposed algorithm achieves an optimal balance between salient anchors (high curvature) and spatial distribution, as demonstrated by the improved reconstruction results in the bottom-right of Figure \ref{fig:anchorComp}.

\section{Limitations}

While our approach demonstrates significant advantages in achieving rotation-equivariant shape completion, it has limitations. Firstly, although our method is less data-driven than techniques that learn rotation through augmentation, it still requires substantial training data to achieve high performance. This dependency on data can be a bottleneck, especially for applications where labeled data is scarce or expensive to obtain.

Another key limitation of our approach is the optimization procedure to find the coordinates of the complete shape. This procedure prevents the model from being rotation-invariant, which will limit its applicability to some real-world applications. 

Furthermore, while distance-based encoding contributes to rotation invariance, it also introduces additional complexity to the learning process. This complexity can sometimes result in lower performance on standard, non-rotated cases. In scenarios where the objects are consistently presented in a canonical alignment, methods that memorize these aligned shapes may outperform our approach. The trade-off between achieving rotation invariance and maintaining high performance on canonical shapes is an essential consideration for the practical deployment of our model.